# Handwritten Bangla Character Recognition Using The State-of-Art Deep Convolutional Neural Networks

**Md Zahangir Alom[1], Peheding Sidike[2], Mahmudul Hasan[3], Tark M. Taha[1], and Vijayan K. Asari[1]**
[1]Department of Electrical and Computer Engineering, University of Dayton, OH, USA
[2]Department of Earth and Atmospheric Sciences, Saint Louis University, St. Louis, MO, USA
[3]Comcast Labs, Washington, DC, USA
Emails: [1]{alomm1, ttaha1, and vasari1}@udayton.edu, [2]sidike.paheding@slu.edu, [3]mahmud.ucr@gmail.com

*Abstract*— **In spite of advances in object recognition technology, Handwritten Bangla Character Recognition (HBCR) remains largely unsolved due to the presence of many ambiguous handwritten characters and excessively cursive Bangla handwritings. Even the best existing recognizers do not lead to satisfactory performance for practical applications related to Bangla character recognition and have much lower performance than those developed for English alpha-numeric characters. To improve the performance of HBCR, we herein present the application of the state-of-the-art Deep Convolutional Neural Networks (DCNN) including VGG Network, All Convolution Network (All-Conv Net), Network in Network (NiN), Residual Network, FractalNet, and DenseNet for HBCR. The deep learning approaches have the advantage of extracting and using feature information, improving the recognition of 2D shapes with a high degree of invariance to translation, scaling and other distortions. We systematically evaluated the performance of DCNN models on publicly available Bangla handwritten character dataset called CMATERdb and achieved the superior recognition accuracy when using DCNN models. This improvement would help in building an automatic HBCR system for practical applications.**

## I. INTRODUCTION

Automatic handwriting character recognition has many academic and commercial interests. Nowadays, Deep Learning techniques already excel in learning to recognize handwritten characters [33]. The main challenge in handwritten character recognition is to deal with the enormous variety of handwriting styles by different writers in different language. Furthermore, some of complex handwriting scripts comprise different styles for writing words. Depending on the language, characters are written isolated from each other in some cases, (*e.g.,* Thai, Laos and Japanese). In some other cases, they are cursive and sometimes the characters are related to each other (*e.g.,* English, Bangladeshi and Arabic). This challenge is already recognized by many researchers in the field of Natural Language Processing (NLP) [1–3]. Handwritten character recognition is more challenging compare to the printed forms of character. In addition, handwritten characters written by different writers are not identical but vary in different aspects such as size and shape. Numerous variations in writing styles of individual character makes the recognition task challenging. The similarities in different character shapes, the overlaps, and the interconnections of the neighboring characters make further complicate the character recognition problem. The large variety of writing styles, writers, and the complex features of the handwritten characters are very challenging for accurately classifying the handwritten characters.

Bangla is one of the most spoken languages and ranked fifth in the world. It is also a significant language with a rich heritage; February 21st is announced as the International Mother Language day by UNESCO to respect the language martyrs for the language in Bangladesh in the year of 1952. This is the only language for which a lot of people sacrifices their life for establishing the Bangla is the first language of Bangladesh and the second most popular language in India. About 220 million people use Bangla as their speaking and writing purpose in their daily life. Therefore, automatic recognition of Bangla characters has a great significance. Different languages have different alphabets or scripts, and hence present different challenges for automatic character recognition respect to language. For instance, Bangla uses a Sanskrit based script which is fundamentally different from English or a Latin-based script. This accuracy for character recognition algorithm may vary significantly depending on the script. Therefore, handwritten Bangla character recognition algorithms should be investigated with due importance.

In Bangla language, there are 10 digits and 50 characters including vowel and consonant, where some contains additional sign up and/or below. Moreover, Bangla consists of many similar shaped characters. In some cases, a character differs from its similar one with a single dot or mark. Furthermore, Bangla language also contains some special characters which equivalent representation of vowels. It makes difficult to achieve a better performance with simple technique as well as hinders to the development of Bangla handwritten character recognition system. There are many applications of Bangla handwritten character recognition such as: Bangla Optical Character Recognition (OCR), National ID number recognition system, automatic license plate recognition system for vehicle and parking lot management system, post office automation, online banking and many more. Some example images are shown in Fig. 1. In this work, we investigate the handwritten character recognition on Bangla numerals, alphabets, and special characters using the state-of-the-art Deep Convolutional

Neural Networks (DCNN). The contributions of this paper are summarized as follows:

- Comprehensive evaluation of the state-of-the-art DCNN models including VGG Net. [19], All-Conv Net. [20], NiN [21], ResNet [22], FractalNet [23], and DenseNet [24] on Bangla handwritten characters recognition.
- Extensive experiments on Bangla handwritten characters recognition including handwritten digits, alphabets and special character recognition
- The best recognition accuracy is achieved compared to many exiting approaches on all experiments.

respectively. However, they did not mention about recognition reliability and response time in their works, which are very important evaluation factors for a practical automatic letter sorting machine. Reliability indicates the relationship between the error rate and the recognition rate. Liu and Suen [10] have shown the benchmarked accuracy of recognition rate of handwritten Bangla digits on a standard dataset, namely the ISI dataset of handwritten Bangla numerals [11], which consists of 19392 training samples, 4000 test samples and 10 classes (i.e., 0 to 9). They have reported accuracy is 99.4% for numeral recognition. Such high accuracy has been attributed to the extracted features based on gradient direction and some

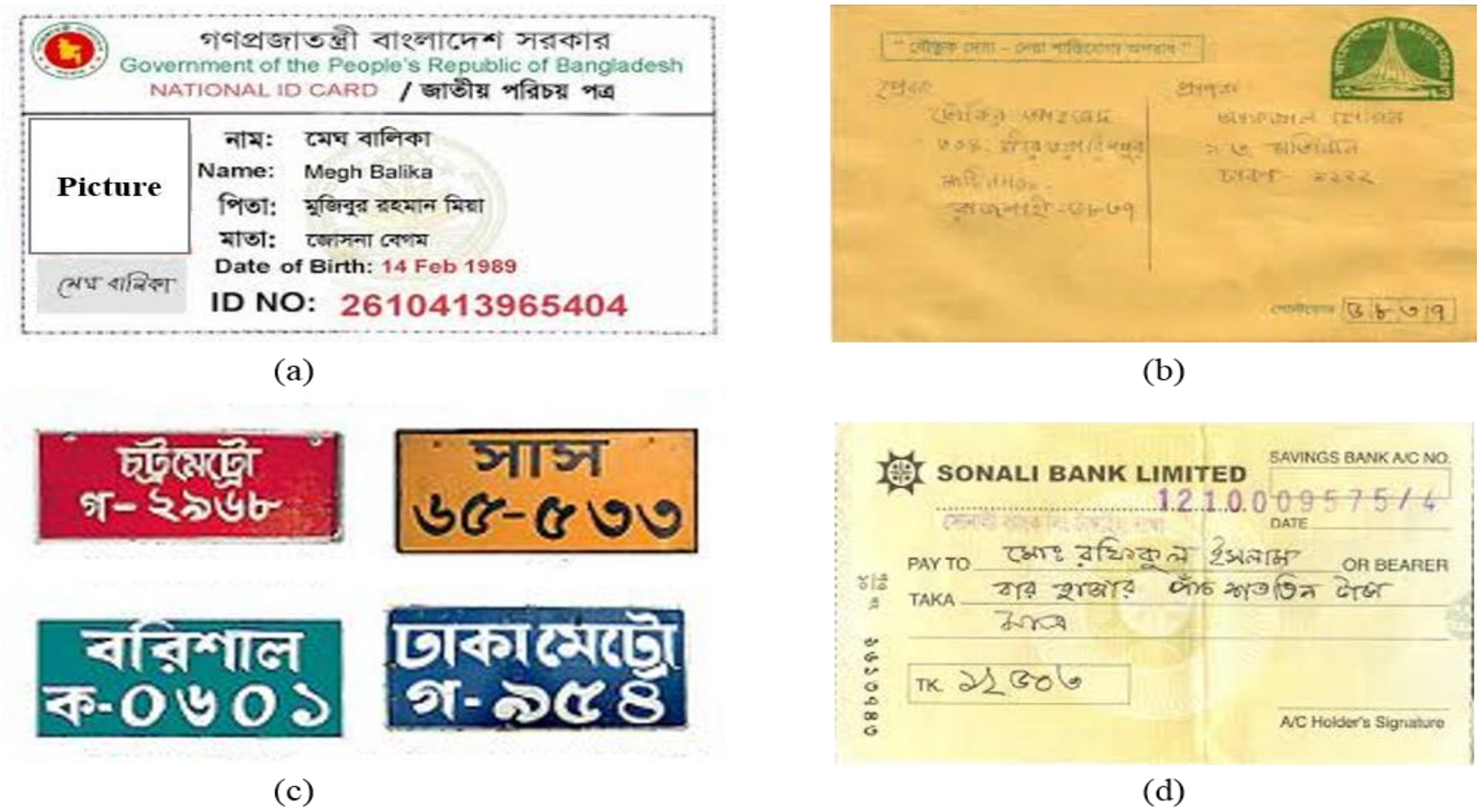

**Fig. 1.** Application of handwritten Character recognition: (a) National ID number recognition system (b) Postal office automation with code number recognition on Envelope (c) Automatic license plate recognition and (d) Bank automation.

The rest of the paper has been organized in the following way: Section II discusses related works. Section III reviews the state-of-the-art DCNNs. Section IV discusses the experimental datasets and results. Finally, the conclusion is made in Section V.

## II. Related Works

There are a few remarkable works are available for Bangla handwritten character recognition. Some literatures have been reported on Bangla characters recognition in the past years [4–6], but there is only few research on handwritten Bangla numeral recognition that reach to the desired recognition accuracy. Pal *et al.* have conducted some exploring works for recognizing handwritten Bangla characters [7–9]. The proposed schemes are mainly based on extracted features from a concept called water reservoir. Reservoir is a concept that obtained by considering accumulation of water poured from the top or from the bottom of the numerals. They deployed a system towards Indian postal automation. The accuracy of the handwritten Bangla and English numeral classifier is 94.13% and 93%, advanced normalization techniques. Surinta *et al.* [12] proposed a system using a set of features such as the contour of the handwritten image computed using 8-directional codes, distance calculated between hotspots and black pixels, and the intensity of pixel space of small blocks. Each of these features is used to a nonlinear SVM classifier separately, and the final decision has been taken based on majority voting. The dataset has used in [12] is composed of 10920 examples, and this method achieves an accuracy of 96.8%. Xu *et al.* [13] used a hierarchical Bayesian network which directly takes raw images as the network inputs and classifies them using a bottom-up approach. Average recognition accuracy of 87.5% was achieved with a dataset consists with 2000 handwritten sample images. Sparse representation classifier is applied for Bangla digit recognition in [14] where 94% accuracy was resulted for handwritten digit recognition. In [15], the handwritten Bangla basic and compound character recognition using MLP and SVM classifier has been proposed and they achieved around 79.73% and 80.9% of recognition rate, respectively. Handwritten Bangla numerals recognition using MLP is presented in [16] where the average recognition rate reached 96.67% using 65 hidden neurons. Das *et al.* [17] exploited genetic algorithms-based region sampling method for local feature selection and achieved 97% accuracy on the handwritten Bangla numeral dataset named CMATERdb. The convolutional neural networks (CNN) based Bangla handwritten character recognition system has been introduced in [18], where the best recognition accuracy is reached at 85.36% on their own dataset for Bangla character recognition. Very recently, deep learning approaches including CNN, CNN with Gabor filters, and Deep Belief Network (DBN) have been applied to handwritten digits recognition [46]. This work has reported the improved recognition accuracy on handwritten Bangla digits recognition. These works lead to the field of deep learning for Bangla character recognition. However, in this paper, we have implemented a set of DCNN including VGG [19], All Conv Net. [20], NiN [21], ResNet [22], FractalNet [23], and DenseNet [24] for Bangla handwritten characters (including digits, alphabets and special characters) recognition. We have achieved the state-of-the-art recognition accuracy in all the mentioned category of Bangla handwritten characters.

## III. Deep Convolutional Neural Networks (DCNN)

In the last few years, deep leaning showed outstanding performance in the field of machine learning and pattern recognition. Deep Neural Networks (DNNs) model generally include Deep Belief Network (DBN) [26, 48], Stacked Auto-Encoder (SAE) [28], and CNN. Due to the composition of many layers, DNN methods are more capable for representing the highly varying nonlinear function compared to shallow learning approaches [25]. The low and middle level of DNNs abstract the feature from the input image whereas the high level performs classification operation on extracted features. As a result, a end-to-end framework is formed by integrating with all necessary modules within a single network. Therefore, DNN models often lead to better accuracy comparing to train each module independently. Among all deep learning approaches, CNN is one of the most popular model and has been providing the state-of-the-art performance on object recognition [49], segmentation [50], human activity analysis [51], image super resolution [52], object detection [53], scene understanding [54], tracking [55], and image captioning [56].

### *A. Convolutional Neural Network (CNN)*

The network model was first time proposed by Fukushima in 1980 [29]. It has not been widely used because the training process was difficult and computationally expensive. In 1998s, LeCun *et al*. applied a gradient-based learning algorithm to CNN and obtained superior performance for digit recognition task [30]. In recent years, there are different variants of new CNN architectures have been proposed for various applications. Cireşan *et al*. applied multi-column CNNs to recognize digits, alpha-numerals, Chinese characters, traffic signs, and object images [31, 32]. They reported excellent results and surpassed conventional methods on many public databases, including MNIST digit image database, NIST SD19 alphanumeric character image dataset, and CASIA Chinese character image datasets. In addition, the advantages of CNNs, CNN approach has been designed to imitate human visual processing, and it has highly optimized structures to process 2D images. Further, CNN can effectively learn the extraction and abstraction of 2D features. The max-pooling layer of CNN is very effective in absorbing shape variations. Moreover, sparse connection with tied weights, CNN has significantly fewer parameters than a fully connected network of similar size. Most of all, CNN is trainable with the gradient-based learning algorithm and suffers less from the diminishing gradient problem. Given that the gradient-based algorithm trains the whole network to minimize an error criterion directly, CNN can produce highly optimized weights. In 2015, Korean or Hangul handwritten character recognition system has been proposed using DCNN and superior performance against classical methods [33].

Fig. 2 shows the overall architecture of the CNN consists with two main parts such as feature extractor and classifier. In the feature extraction unit, each layer of the network receives the output from its immediate previous layer as inputs and passes current output as inputs to the immediate next layer. The basic CNN architecture is consisted with the combination of three types of layers: convolution, max-pooling, and classification [30]. There are two types of layers in the low and middle-level of the network such as convolutional layer and max-pooling layers. In the very basic architecture, the even numbered layers work for convolution and odd numbered layers work for max-pooling operation. The output nodes of the convolution and max-pooling layers are grouped into 2D planes which is called feature mapping. Each plane of the layer usually derived with the combination of one or more planes of the previous layers. The node of the plane is connected to a small region of each connected planes of the previous layer. Each node of the convolution layer extracts the features from the input images by convolution operation on the input nodes. The max-pooling layer abstracts the feature through average or propagating the operation on input nodes. The higher-level features can be derived from the propagated feature of the lower level layers. As the features propagate to the highest layer, the dimension of the feature is reduced depending on the size of the convolutional and max-pooling masks. However, the number of feature mapping usually increased for selecting or mapping the extreme suitable features of the input images for better classification accuracy. The outputs of the last layer of CNN are used as inputs to the fully connected network, which is called classification layer [30].

Finally, we used a Softmax, or normalized exponential function layer at the end of the architecture. The score of the respective class has been calculated in the top classification layer through propagation. Based on the highest score, the classifier gives outputs for the corresponding classes after completing the propagation. For an input sample *x*, weight vector *W*, and *K* distinct linear functions, the Softmax operation can be defined for the $i^{th}$class as follows:

$$P(y=i|x)=\frac{e^{x^T w_i}}{\sum_{k=1}^{K} e^{x^T w_k}} \tag{1}$$

However, there are different variant of DCNN architecture have been proposed last few years. The following section discusses on different modern DCNN models.

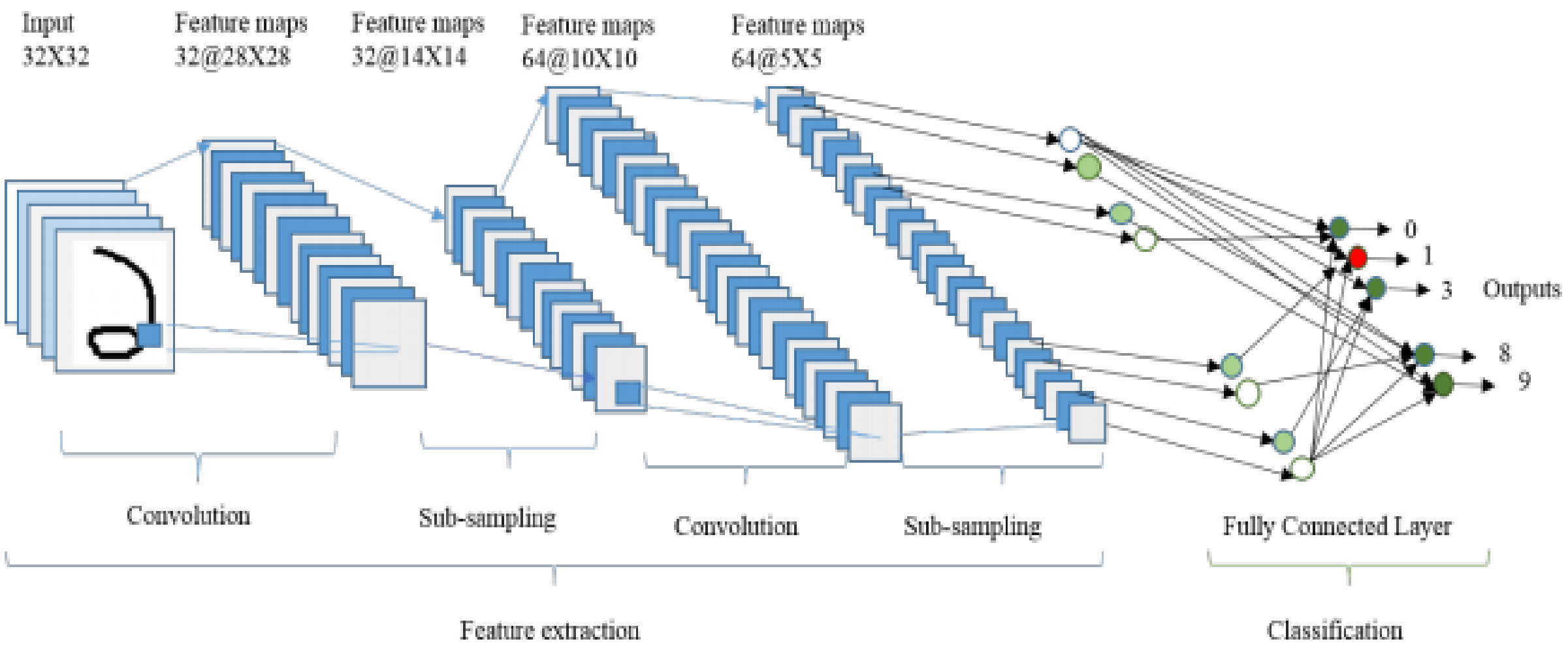

**Fig. 2.** Basic CNN architecture for digit recognition

## B. *CNN Variants*

As far as CNN architecture is concerned, it can be observed that there are some important and fundamental components that are used to construct an efficient DCNN architecture. These components are: convolution layer, pooling layer, fully-connected layer, and Softmax layer. The advanced architecture of this network consists of stack of convolutional layers and max-pooling layers followed by fully connected and softmax layer at the end. Noticeable examples of such networks include LeNet [30], AlexNet [36], VGG Net [19], NiN [21] and All Convolutional (All Conv) [20]. There are some other alternative and efficient advanced architectures have been proposed including GoogleNet with Inception layers [37], Residual Network [22], FractalNet [23] and DenseNet [24]. However, there are some topological difference is observed in the modern architectures. Out of many DCNN architectures, AlexNet [36], VGG Net [19], Google Net [37], Residual Network [22], Dense CNN called DenseNet [24] and FractalNet [23] can be premeditated most popular architectures with respect to their enormous performance on different benchmarks for object classification. Among these models, some of the models are designed especially for large scale implementation such as ResNet and GoogleNet, whereas the VGG Net consists of a general architecture. Some of the architectures are very dense in term of connectivity like DenseNet. On the other hand, Fractal Network is an alternative of ResNet. In this paper, we have implemented All-Conv, NiN, VGG-16, ResNet, FractralNet, and DenseNet. The basic overview of these architectures is given in the following section.

### 1) VGG-16 NET

The Visual Geometry Group (VGG) was the runner up of ImageNet Large Scale Visual Recognition Competition (ILSVRC) in 2014 [9]. The main contribution of this model shows that the depth (i.e., number of layers) of a convolutional neural network is the critical component of high recognition or classification accuracy. In this architecture, two convolutional layers are used consecutively with a rectified linear unit (ReLU) activation function followed by single max-pooling layer, several fully connected layers with ReLU and soft-max as the final layer. The 3×3 convolutional filters with stride 2 is applied for performing filtering and sub-sampling operations simultaneously in VGG-E version [19]. There are three types of VGGNet based on the architecture. These three network contains 11, 16 and 19 layers and named as VGG-11, VGG-16 and VGG-19, respectively. The basic structure for VGG-11 architecture is shown in Fig. 3. There are five convolution layers, three max-pooling layers and three fully connected (FC) layers.

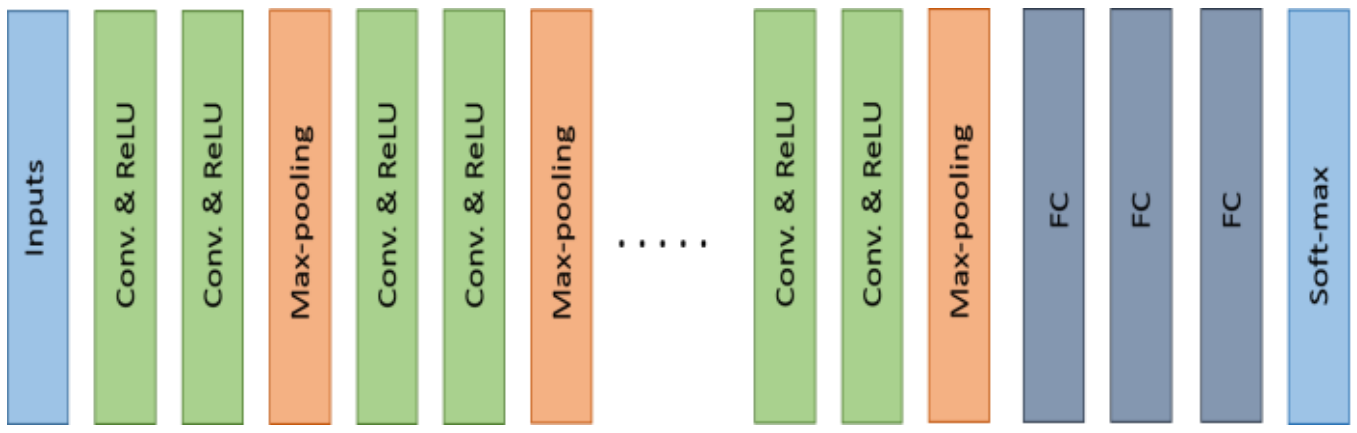

**Fig. 3.** Basic architecture of VGG Net.: Convolution (Conv) and FC for fully connected layers and Softmax layer at the end.

The configuration of VGG16 is as follows: number of convolution layers: 16, fully connected layers: 3, weights: 138 Million and Multiplication and Accumulates (MACs): 15.5G. In this architecture, the multiple convolutional layers are incorporated which is followed by a max-pooling layer. In this implementation, we have used VGG16 with less number of feature maps in each convolutional block compared to standard VFF16 model.

### 2) *All Convolutional Network (All-Conv)*

The basic layer specification of All-Conv Network is given in Fig. 4. This basic architecture is composed with two convolution layers followed by a max-pooling layer. Instead of using fully connected layer, Global Average Pooling (GAP) with the dimension of 6×6 is used. Finally, the Softmax layer is used for classification. The output dimension is assigned based on the number of classes.

| |
|---|
| Input 32×32 RGB image |
| 3×3 Conv. 128 ReLU<br>3×3 Conv. 128 ReLU |
| 3×3 Max-pooling stride 2 |
| 3×3 Conv. 256 ReLU<br>3×3 Conv. 256 ReLU |
| 3×3 Max-pooling stride 2 |
| 3×3 Conv. 512 ReLU |
| 1×1 Conv. 512 ReLU |
| Global Average over 6×6 spatial dimensions |
| 10/50/13 - way softmax |

**Fig. 4.** Model of All Convolutional Network

### 3) *Network in Network (NiN)*

This model is quite different compared to the above mentioned DCNN models due to the following reasons [21]:

1. It uses multilayer convolution where Convolution is performed with 1×1 filters.
2. This model uses Global Average Pooling (GAP).

The concept of using 1×1 convolution helps to increase the depth of the network which is regularized by dropout method. The GAP significantly change the network structure which is used nowadays very often as a replacement of fully connected layers. The GAP on a large feature map is used to generate a final low dimensional feature vector instead of reducing the feature map to a small size and then flattening the feature vector.

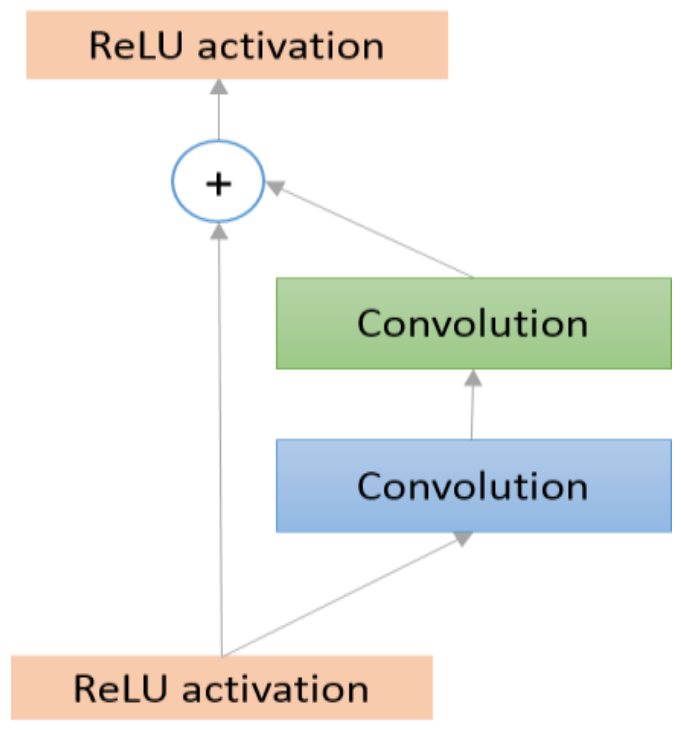

**Fig. 5.** Basic diagram of Residual block

### 4) *Residual Network (ResNet)*

Residual Network developed by He et al. and won the ILSVRC award in 2015 [22]. Nowadays, this new network architecture becomes very popular in computer vision and deep learning community [37]. The proposed version has been experimented with different number of layers. The configuration is summarized as follows: number of convolution layers: 49 (34, 152, 1202 layers for other versions of ResNet), number of fully connected layers: 1, weights: 25.5M and MACs: 3.9G. The basic block diagram of ResNet architecture is shown in Fig. 5. ResNet is a traditional feed forward network. Let's considered the input of the residual block is $x_{l-1}$ and output of this block is $x_l$. After performing operations (e.g. convolution with different size of filters, batch normalization (BN) followed by activation function such ReLU) on $x_{l-1}$, the output $\mathcal{F}(x_{l-1})$ is produced. The final output of the residual unit is defined as follows:

$$x_l = \mathcal{F}(x_{l-1}) + x_{l-1} \qquad (2)$$

The residual network consists of several basic residual units. The different residual units are proposed with different types of layers. However, the operation between the residual units vary depending on the architectures which are explained in [22].

### 5) *FractalNet*

The FractalNet architecture is an advanced and alternative one of ResNet, which is very efficient for designing very large network with nominal depth, but shorter paths for the propagation of gradient during training [23]. This concept is based on drop-path which is another regularization for large network. As a result, this concept helps to enforcing speed versus accuracy tradeoff. The basic block diagram of FractalNet is shown in Fig. 6. Here *x* is the actual inputs of FractalNet, *z* and *f(z)* are the inputs and outputs of Fractal-block respectively.

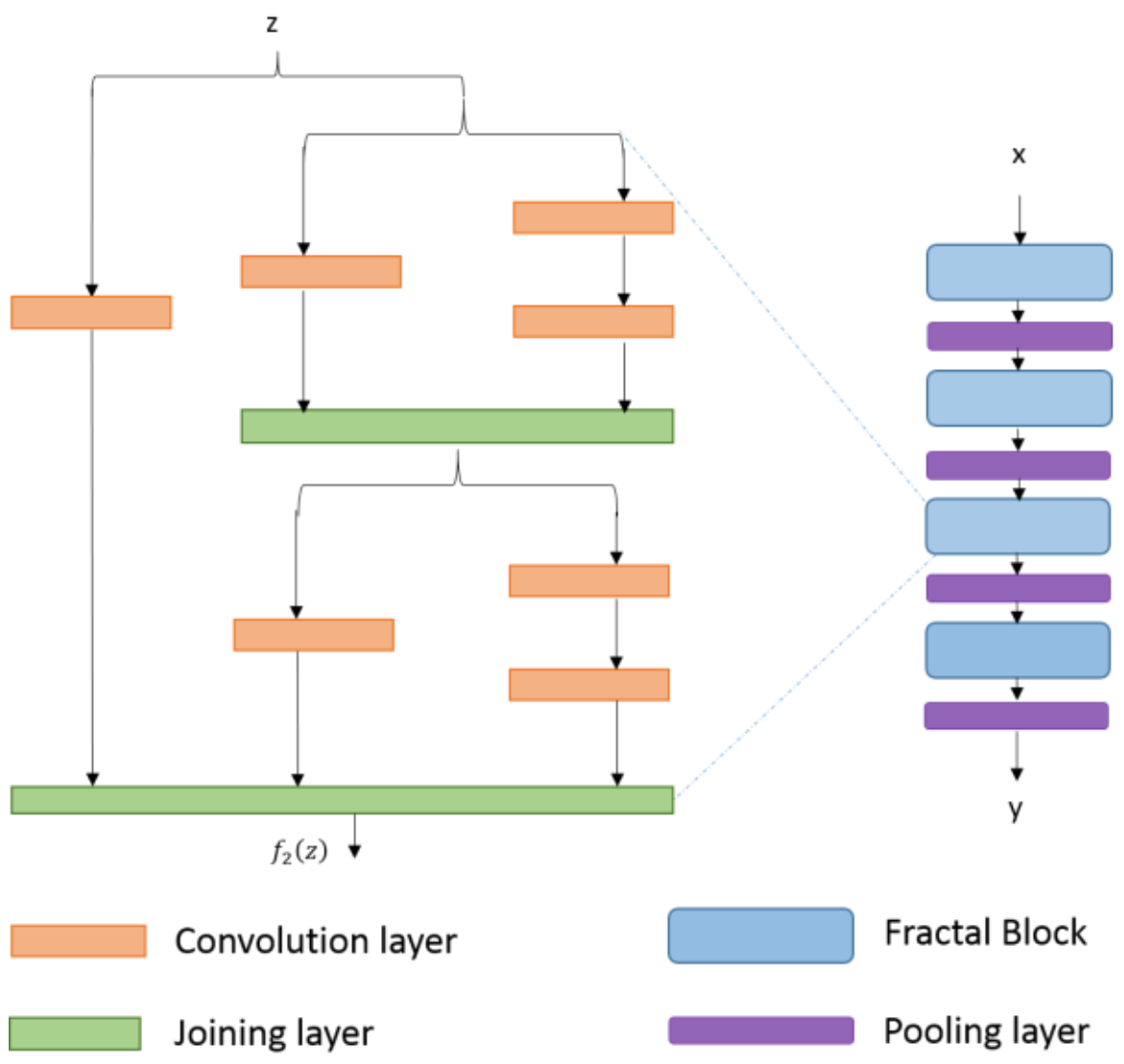

**Fig. 6.** FractalNet module on the left and FractalNet on the right

### 6) *Densely Connected Network (DenseNet)*

DenseNet was developed by Huang et al., which is densely connected CNN where each layer is connected to all previous layers [24]. Therefore, it forms very dense connectivity between the layers and so it is called "DenseNet". The DenseNet consists of several dense blocks, and the layer between two adjacent blocks is called transition layers. The conceptual diagram of the dense block is shown in Fig. 7. According to the figure, the $l^{th}$ layer receive all the feature maps $(x_0, x_1, x_2 \cdots x_{l-1})$ from the previous layers as input:

$$x_l = H_l([x_0, x_1, x_2 \cdots x_{l-1}]) \qquad (4)$$

where $[x_0, x_1, x_2 \cdots x_{l-1}]$ is the concatenated features from $0, \cdots\cdots, l-1$ layers and $H_l(\cdot)$ is a single tensor. It performs three consecutive operation: BN [57], followed by ReLU and a $3 \times 3$ convolution (conv). In the transition block, $1 \times 1$ convolutional operations are performed with BN followed by $2 \times 2$ average pooling layer. This new architecture has achieved state-of-the-art accuracy for object recognition on the five different competitive benchmarks.

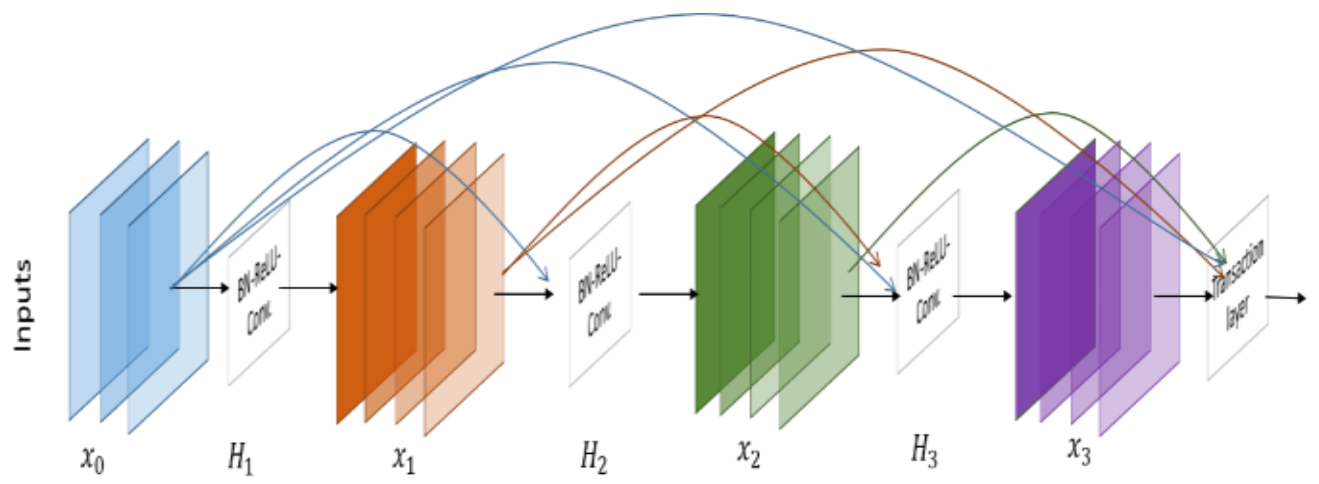

**Fig. 7.** A 4-layerDense block with growth rate of $k = 3$. Each of the layer takes all of the preceding feature maps as input.

### *C. Network parameters*

The number of network parameters is very important criteria to assess the complexity of the architecture. The number of parameters can be used to make comparison between different architectures, calculated as follows. At first the dimension of the output feature map can be compute as,

$$M = \frac{(N - F)}{S} + 1 \tag{5}$$

where $N$ refers the dimension of input feature maps, $F$ refers the dimension of filters or receptive field, $S$ refers to stride in the convolution, and $M$ refers the dimension of output feature maps. The number of network parameters for a single layer is obtained by

$$P_l = (F \times F \times FM_{l-1}) \times FM_l \tag{6}$$

where $P_l$ represents the total number of parameters in the $l^{th}$ layer, $FM_l$ is the total number of output feature maps of $l^{th}$ layer and $FM_{l-1}$ is the total number of input feature maps of $(l-1)^{th}$ layer. For example, let's consider a 32×32 dimensional (N) images as input. The size of the filter (F) is 5×5 and stride (S) is 1 for convolutional layer. The output dimension (M) of the convolutional layer is 28×28 which is calculated according to Eq. 5, if we consider six output mapping in the convolutional layer.

$$P_l = (F \times (F + 1)) \times FM_{l-1}) \times FM_l \tag{7}$$

The number of parameters with bias are used to learn is ((5×5+1) × 3) × 6 = 540 according to the Eq. 7. For 6 output feature maps, the total number of connections are 28×28× (5×5+1) ×6=122,304. For the subsampling layer, the nubmer of trainable parameters is 0. A summary of parameters of All Convolutional architecture are shown in Table I.

## IV. Experimental Results And Discussion

The entire experiment has been conducted on Desktop computer with Intel® Core-I7 CPU @ 3.33 GHz, 56.00GB memory, and Keras with Theano on the backend on Linux environment. We evaluate the state-of-the-art DCNN models on three datasets for Bangla handwritten digits, alphabets, and special characters recognition.

**Table I.** The total number of parameters for All Convolutional Network.

| **Layers** | **Operations** | **Feature maps** | **Size of feature maps** | **Size of kernels** | **# parameters** |
|---|---|---|---|---|---|
| Inputs | | 32× 32× 3 | | | |
| $C_1$ | Convolution | 128 | 30×30 | 3×3 | 3,456 |
| $C_2$ | Convolution | 128 | 28×28 | 3×3 | 147,456 |
| $S_1$ | Max-pooling | 128 | 14×14 | 2×2 | N/A |
| $C_3$ | Convolution | 256 | 12×12 | 3×3 | 294,912 |
| $C_4$ | Convolution | 256 | 10×10 | 3×3 | 589,824 |
| $S_2$ | Max-pooling | 256 | 5×5 | 2×2 | N/A |
| $C_5$ | Convolution | 512 | 3×3 | 3×3 | 1,179,648 |
| $C_6$ | Convolution | 512 | 3×3 | 1×1 | 26,624 |
| $GAP_1$ | GAP | 512 | 3×3 | N/A | N/A |
| Outputs | Softmax | 10 | 1x1 | N/A | 5,120 |

The statistics of three datasets are summarized in Table II. For our convenience, we name the datasets as Digit-10, Alphabet-50, and SpecialChar-13, respectively. We have rescaled all the images to 32×32 pixels for this experiment.

**Table II.** Database statistics used in our experiment

| **Dataset** | **Number of training samples** | **Number of testing samples** | **Total samples** | **Number of classes** |
|---|---|---|---|---|
| Digit-10 | 4000 | 2000 | 6000 | 10 |
| Alphabet-50 | 12,000 | 3,000 | 15,000 | 50 |
| SpecialChar-13 | 2196 | 935 | 2231 | 13 |

### *A. Bangla Handwritten digit dataset*

We evaluate the performance of both DBN and CNN on a Bangla handwritten benchmark dataset called CMATERdb 3.1.1 [44].

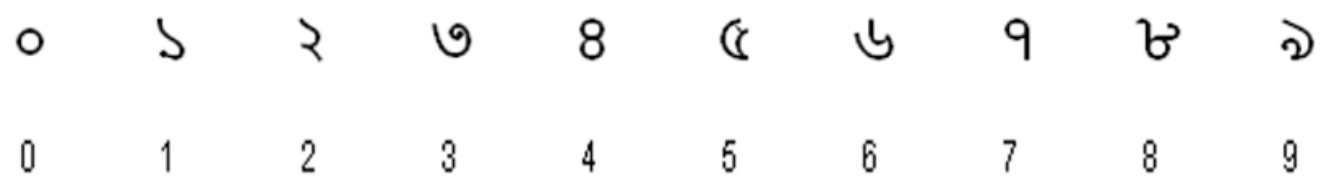

**Fig. 8**. Bangla actual digits in the first row and second row shows the corresponding English digits.

The standard samples of the numeral with respective English numeral are shown in Fig. 8 This dataset contains 6000 images of unconstrained handwritten isolated Bangla numerals. Each digit has 600 images that are rescaled to 32×32 pixels. Some sample images of the database are shown in Fig. 9. Visual inspection depicts that there is no visible noise. However, variability in writing style due to user dependency is quite high. The data set was split into a training set and a test set for the evaluation of different DCNN models. The training set consists of 4000 images (400 randomly selected images of each digit). The rest of the 2000 images are used for testing [46].

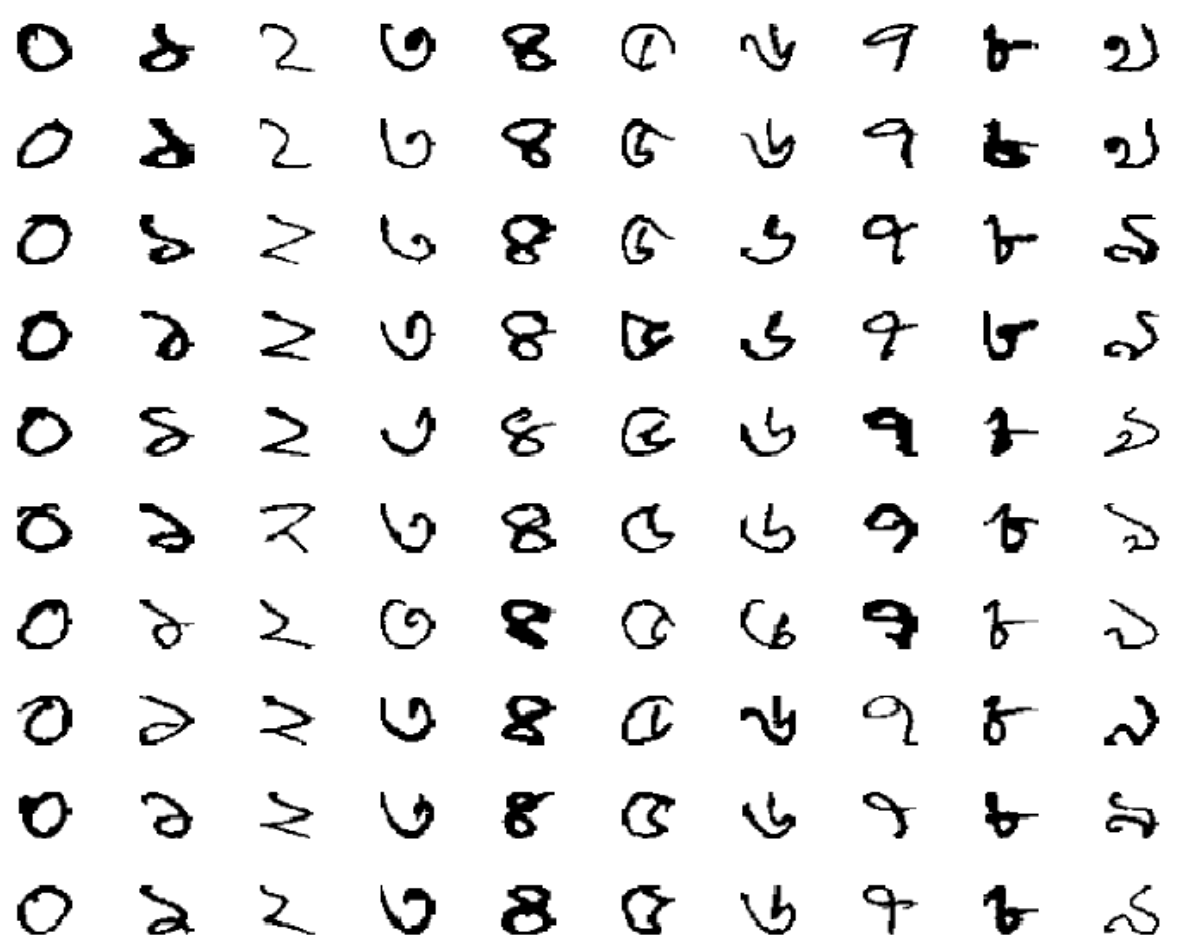

**Fig. 9.** Sample handwritten Bangla numeral images: 1-10 illustrate some randomly selected handwritten Bangla numeral images with

Figure 10 shows the training loss of all DCNN models for 250 epochs. It can be observed that FractalNet and DenseNet converge faster compared to other networks, and worst convergence is obtained to be for the All-Conv network.

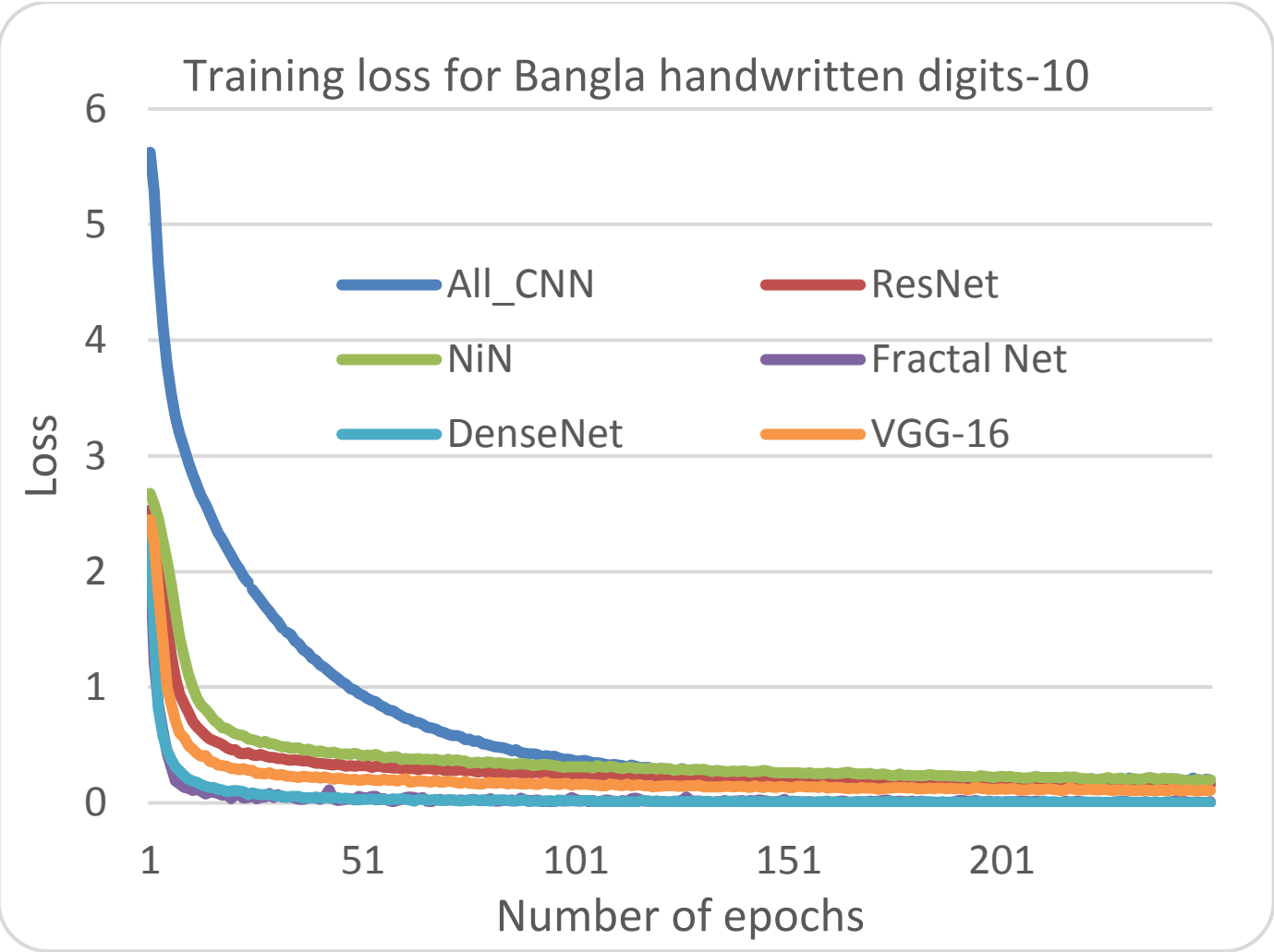

**Fig. 10** Training loss of different architecture for Bangla handwritten 10 digits

The validation accuracy is shown in Fig. 11 where DenseNet and FractalNet show better recognition accuracy among all DCNN models.

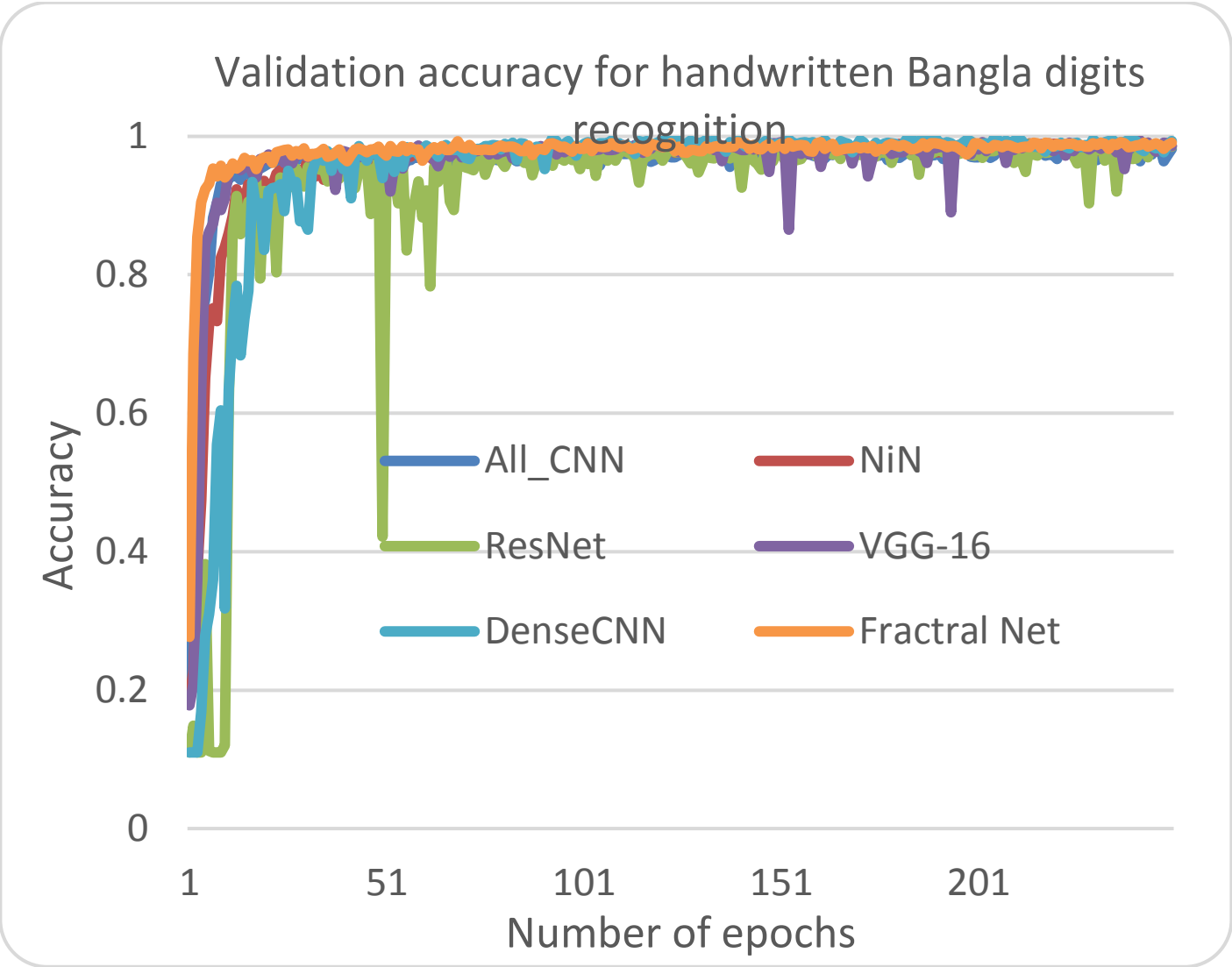

**Fig. 11.** Validation accuracy of different architectures for Bangla handwritten 10 digits.

The testing accuracy of all the DCNN models are shown in Fig. 12. From the result, it can be clearly seen that DenseNet provides the best recognition accuracy compared to other networks.

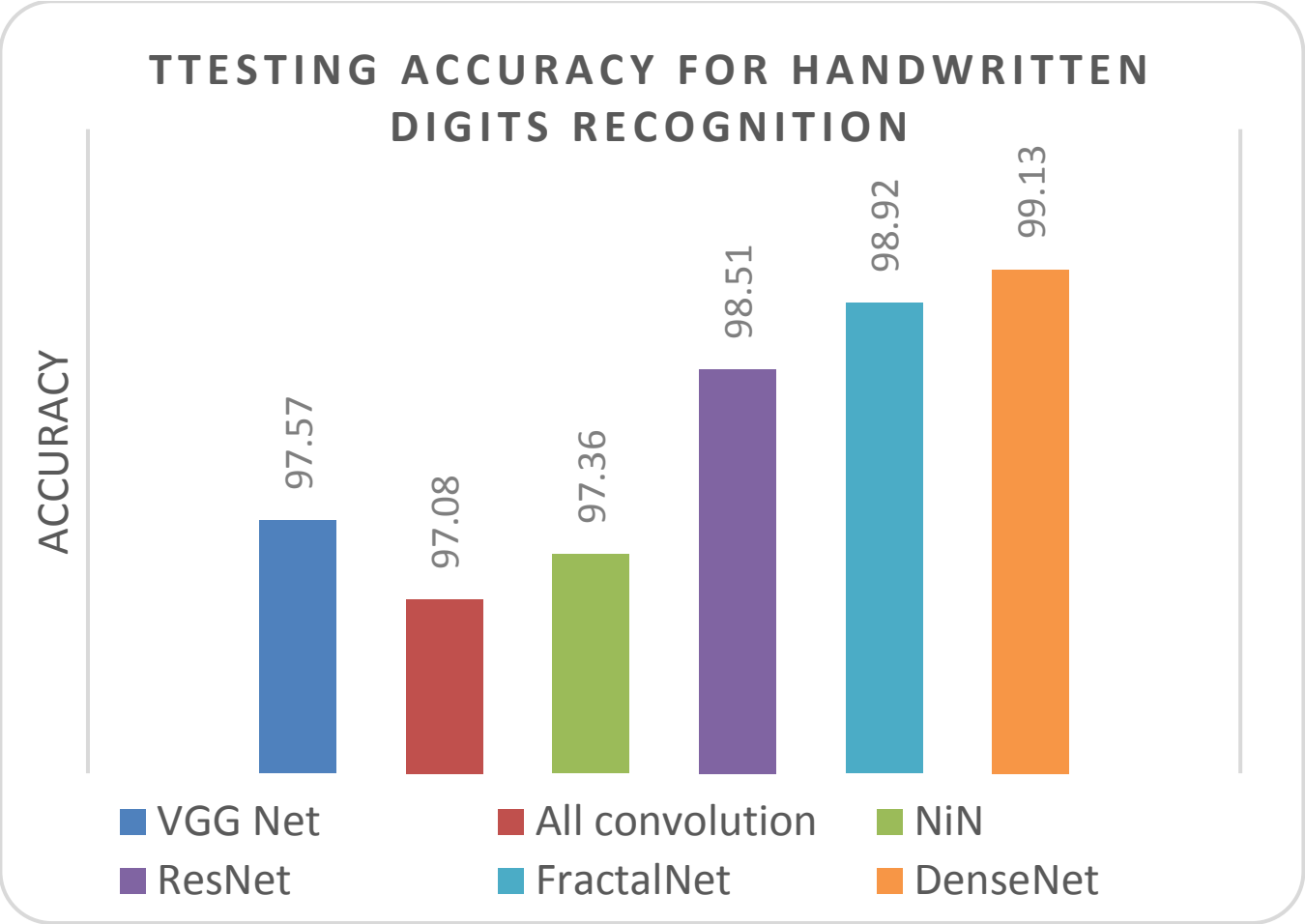

**Fig. 12.** Testing accuracy for Bangla handwritten digits recognition

## B. *Bangla Handwritten 50-alphabet*

In our implementation, the basic fifty alphabets including 11 vowels and 39 consonants are considered. The samples of 39-consonant and 11-vowel characters are shown in Fig. 13(a) and (b) respectively.

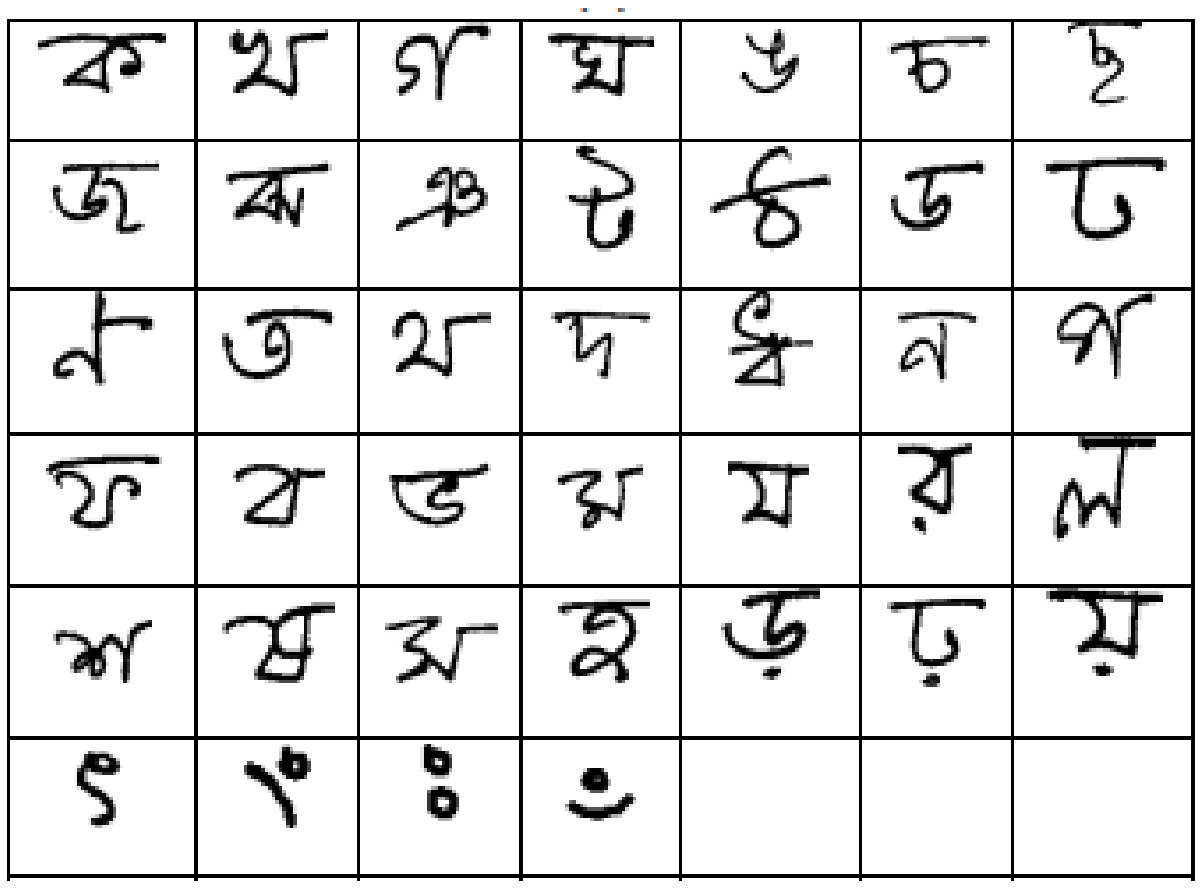

(a)

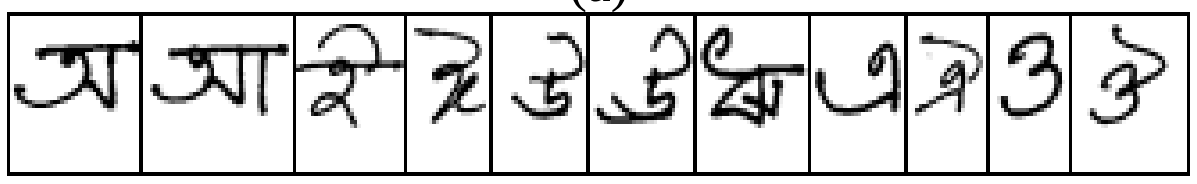

(b)

**Fig. 13.** Example images of handwritten characters: (a) Bangla consonants Characters and (b) vowels.

This dataset contains 15,000 samples where 12,000 are used for training and the remaining 3000 samples are used for testing. The dataset contains samples with different dimension, we rescale all input images to 32×32 pixels. The randomly selected samples from this database are shown in Fig. 14.

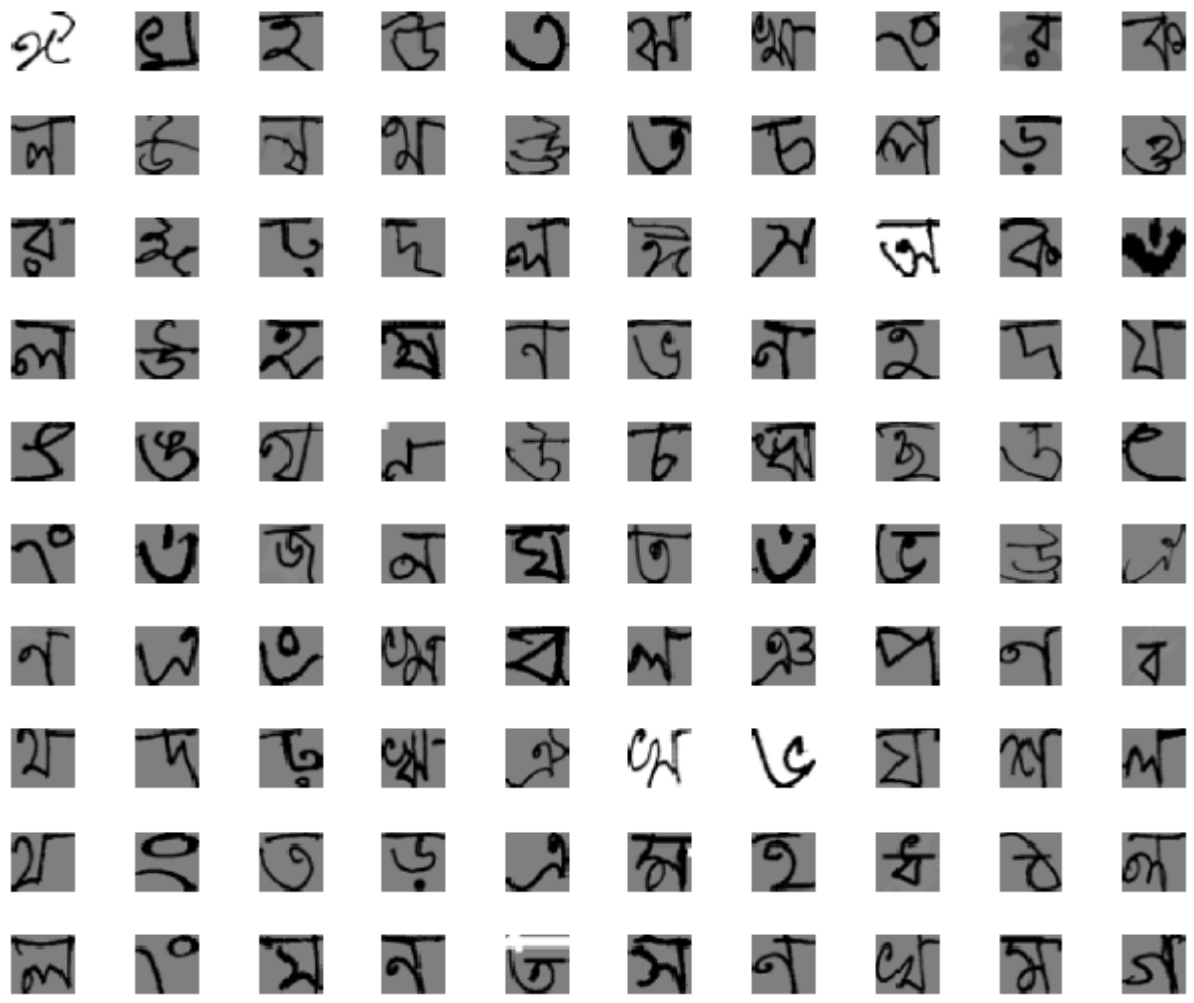

**Fig. 14.** Randomly selected handwritten characters of Bangla Alphabets from dataset.

The training loss for different DCNN models are shown in Fig. 15. It is cleared that the DenseNet shows the best convergence comparing to the other DCNN approaches. Like the previous experiment the All Conv Net shows the worst convergence behavior. In addition, an unexpected convergence behavior is observed in the case of NiN model. However, all DCCN models tend to converge after 200 epochs.

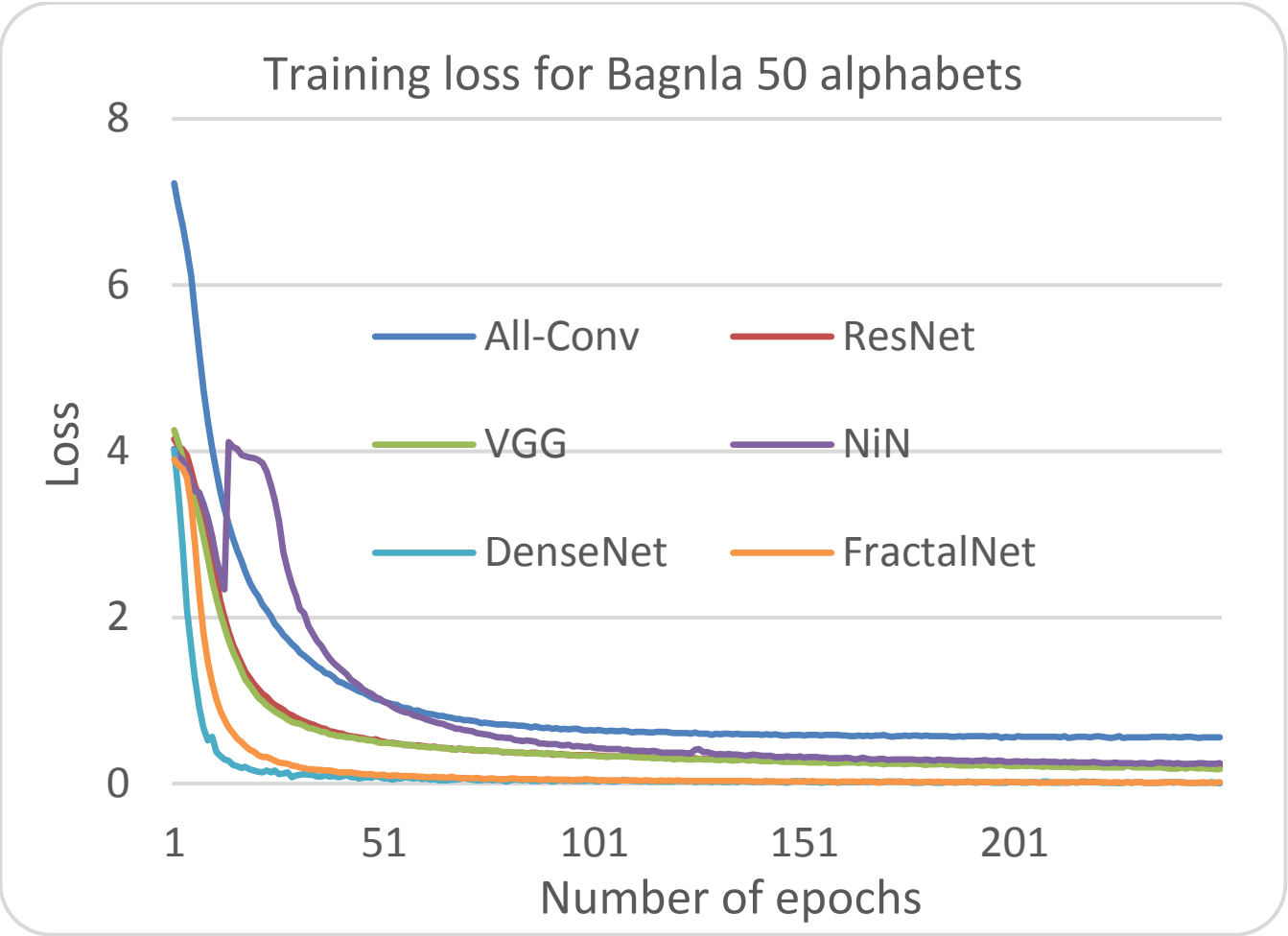

**Fig. 15.** Training loss of different DCNN models for Bangla handwritten 50-alphabets.

The validation accuracy on Alphabet-50 is shown in Fig. 16. DenseNet again shows superior validation accuracy compared to other DCNN approaches.

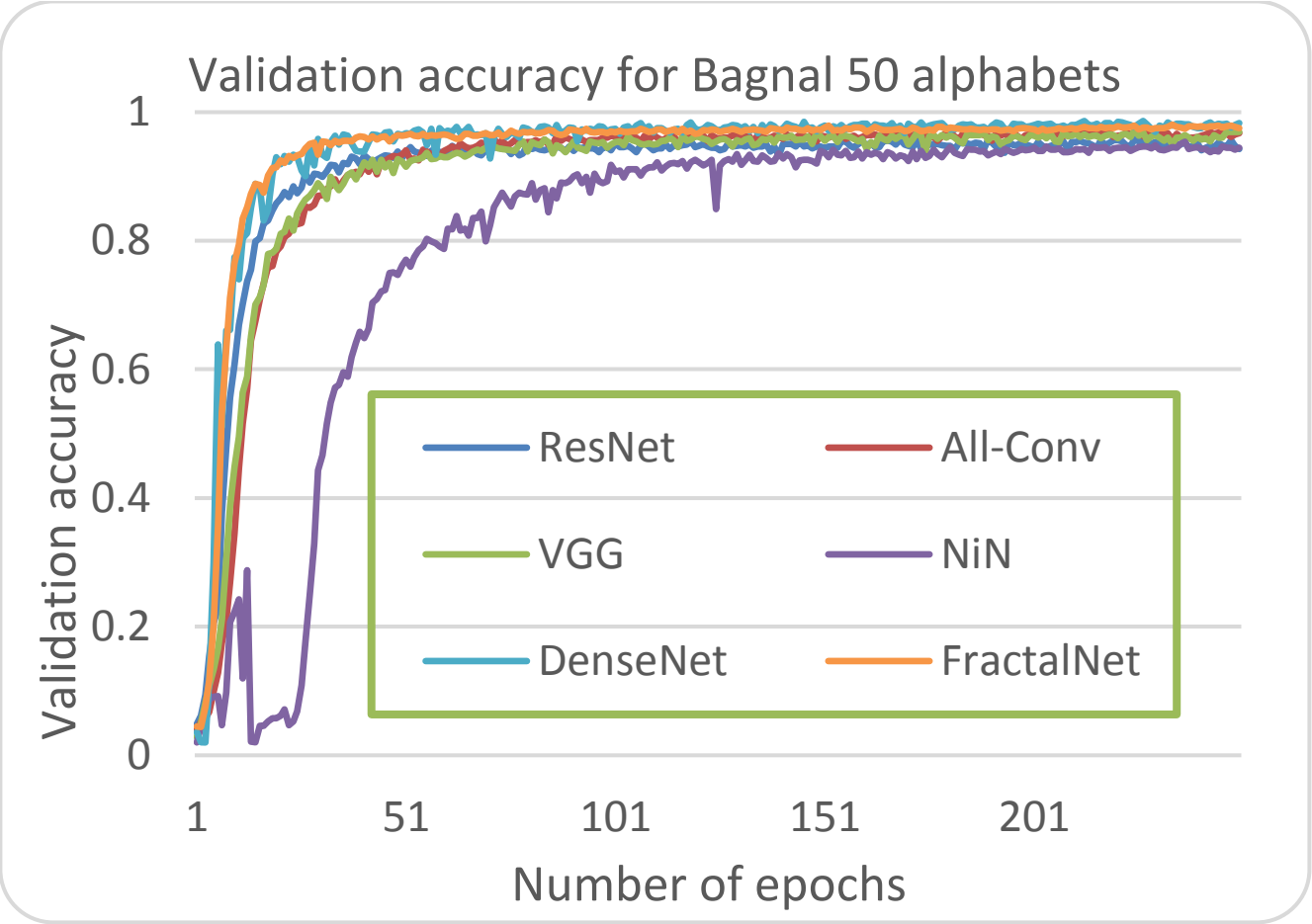

**Fig. 16.** The validation accuracy of different architecture for Bangla handwritten 50-alphabet.

The following bar graph shows the testing results on hand-written Alpabet-50. The DenseNet shows the best testing accuracy with a recognition rate of 98.31%. On the other hand, the All Conv Net provides around 94.31% testing accuracy which is the lowest testing accuracy among all the DCNN models.

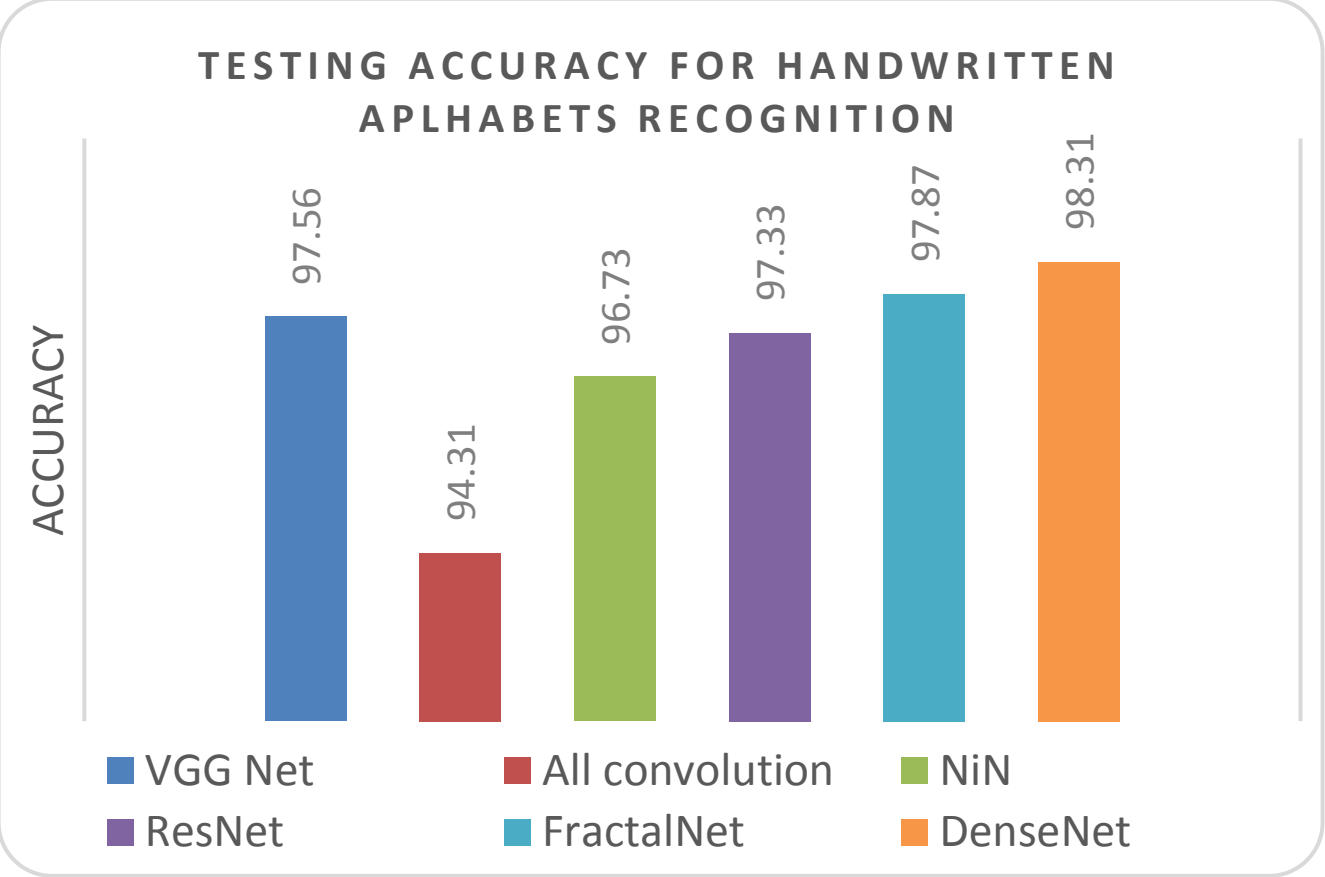

**Fig. 17.** Testing accuracy for handwritten 50-alphabets recognition using different DCNN techniques

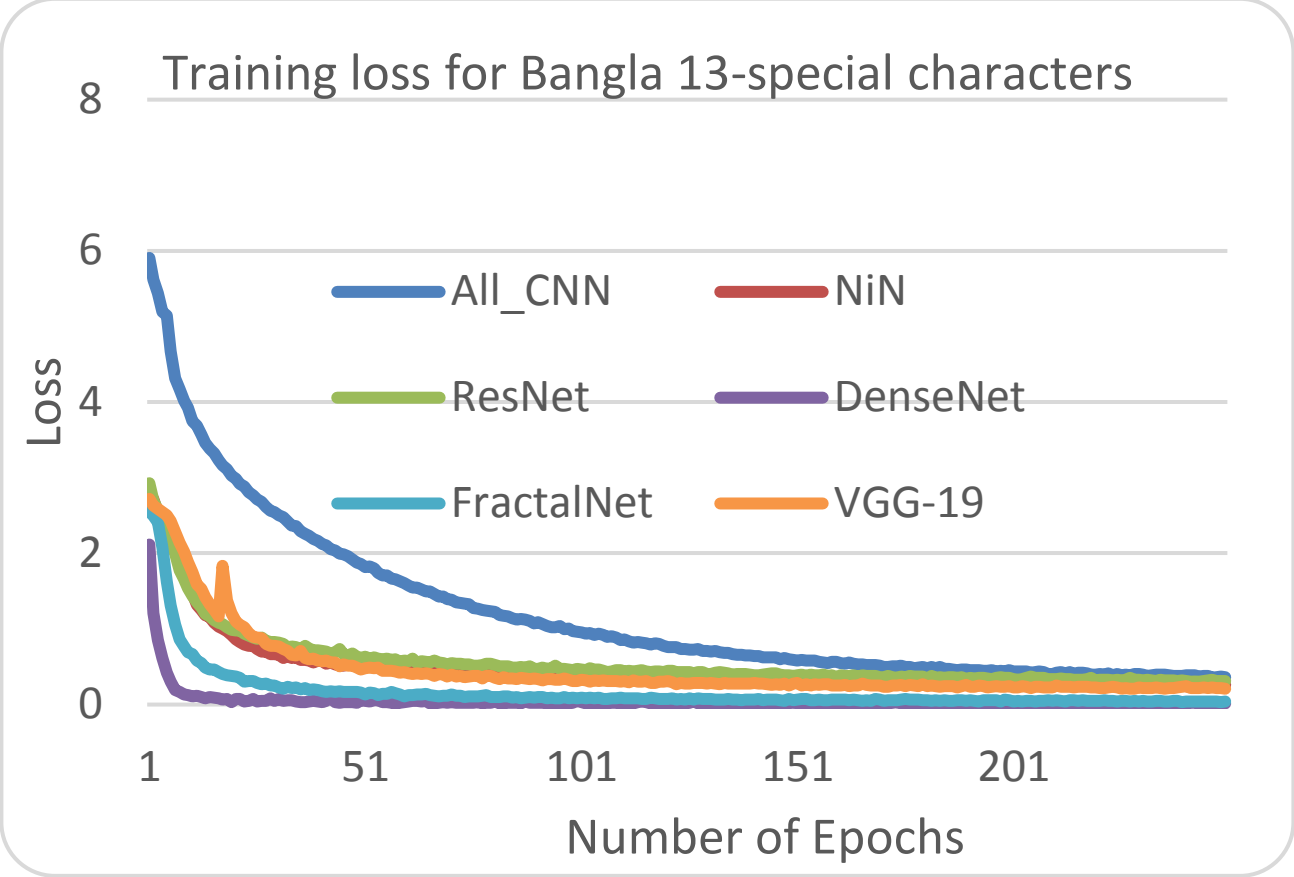

**Fig. 19.** Training loss of different architecture for Bangla 13 special characters (SpecialChar-13)

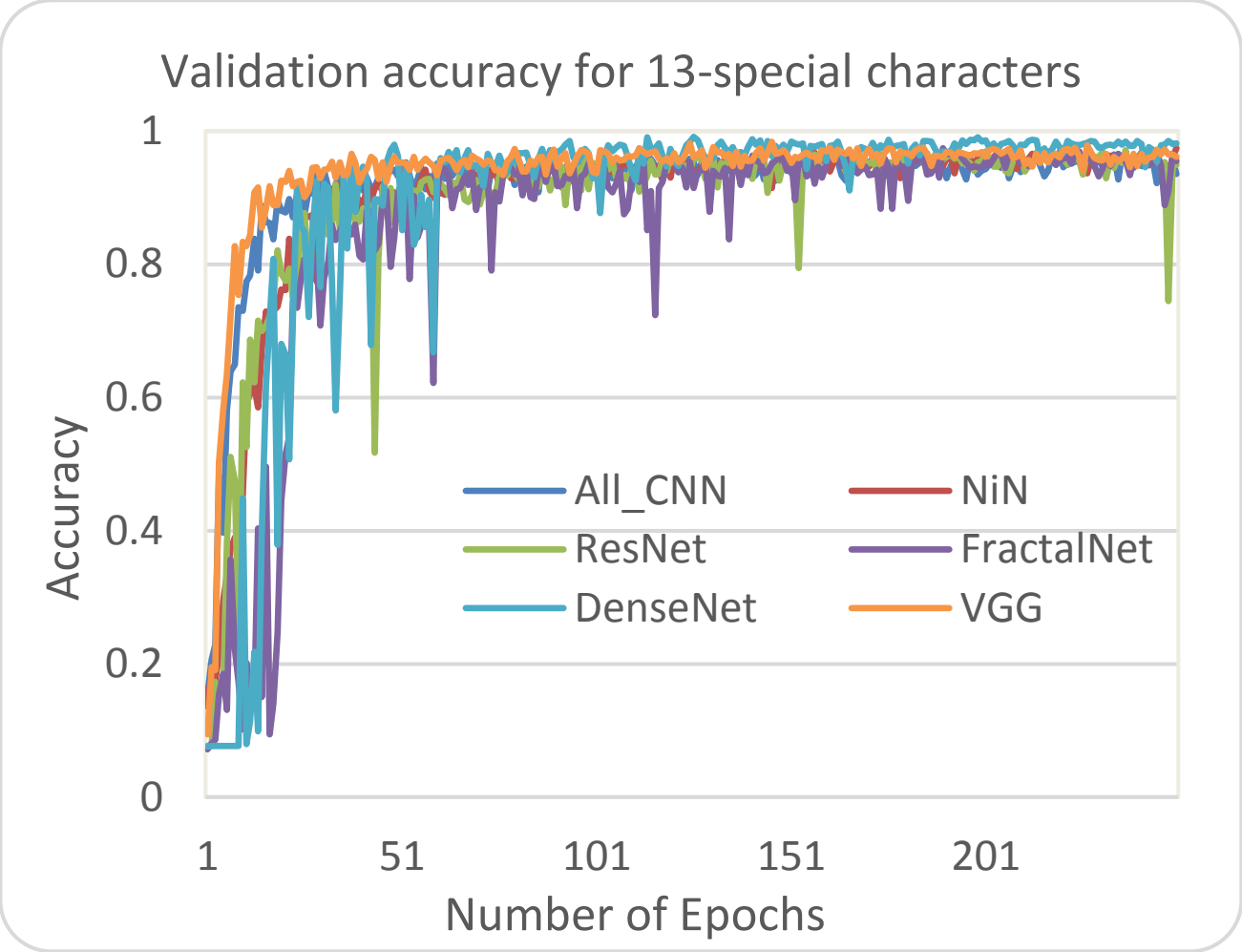

**Fig. 20.** Validation accuracy of different architecture for Bangla 13 special characters (SpecialChar-13)

## C. *Bangla Handwritten Special Characters*

There are several special characters (SpecialChar-13) which is equivalent representation of vowels that are combined with consonants for making meaningful words. In our evaluation, we used 11 special characters which are for 11 vowels and two additional special characters. Some samples of Bangla special characters are shown in Fig. 18. It can be seen from the figure that the quality of the samples is not good, and different variants of writing of the same symbols make this recognition task even difficult.

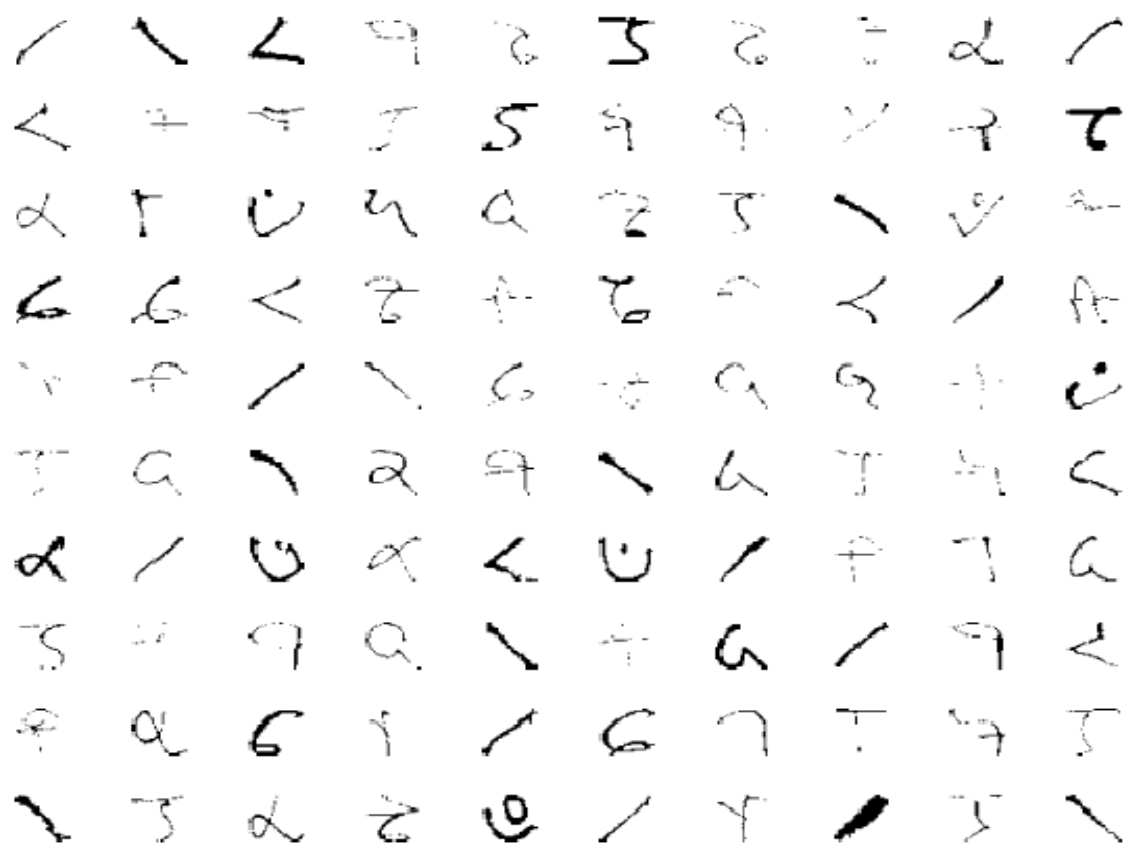

**Fig. 18.** Randomly selected images of special character from dataset.

The training loss and validation accuracy for SpecialChar-13 is shown in Fig. 19 and Fig. 20 respectively. From these figures, it can be said that the DenseNet provides better performance with lower loss and with the highest validation accuracy among all DCNN models. Fig. 21 shows the testing accuracy of DCNN model for SpecialChar-13 dataset. It is observed from the Fig. 21 that DenseNet show highest testing accuracy with lowest training loss and it converges very fast. On the other hand, VGG-19 network shows promising recognition accuracy as well.

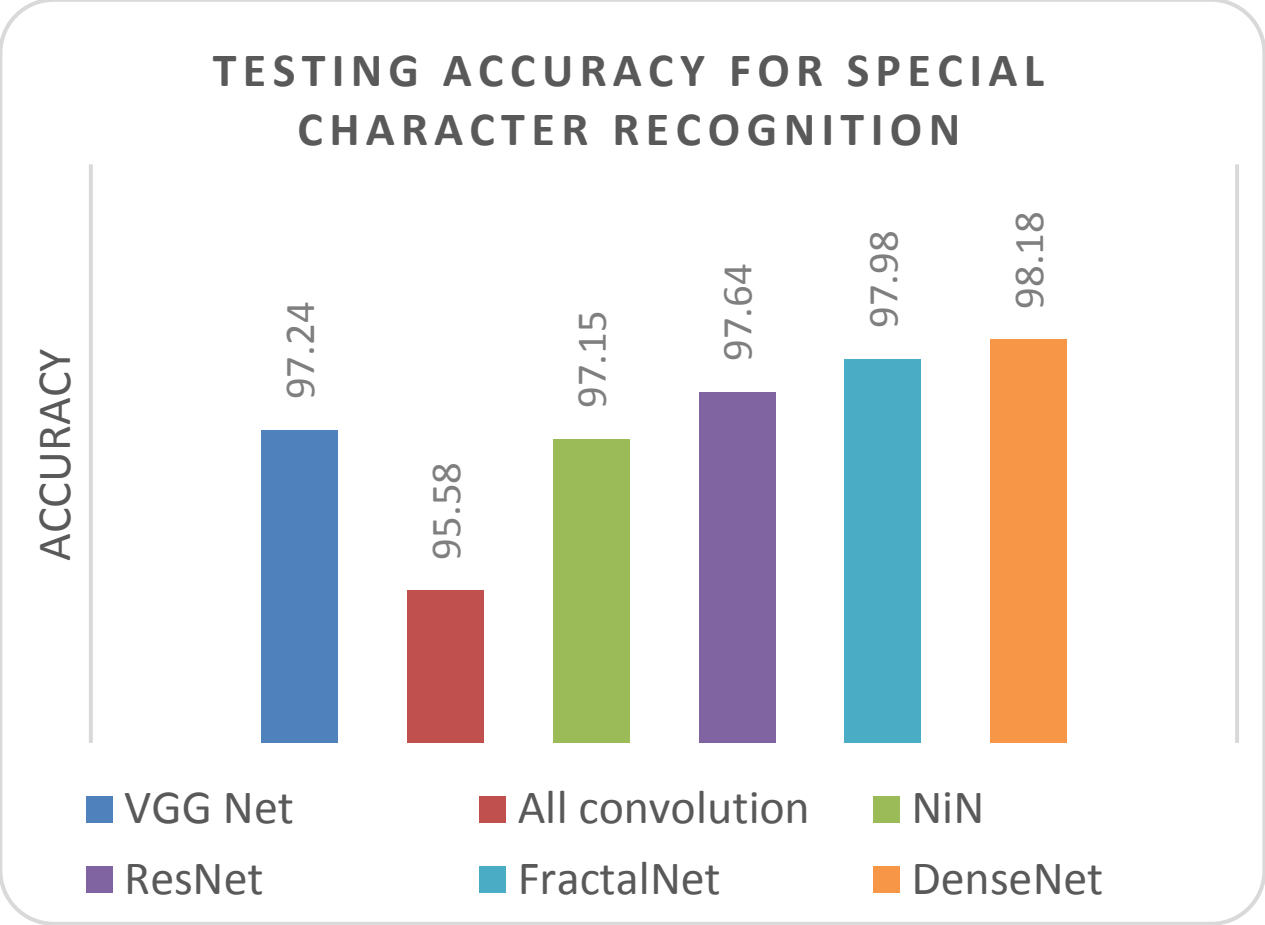

**Fig. 21.** Testing accuracy of different architecture for Bangla 13 special characters (SpecialChar-13)

**Table III.** The testing accuracy of VGG-16 Network, All Conv. Network, NiN, ResNet, FractalNet, and DenseNet on Digit-10, Alphabet-50 and SpecialChar-13 and comparison against other exiting methods.

| Types | Method name | Accuracy of Digit-10 | Accuracy of Alphabet-50 | Accuracy of SpecialChar-13 |
|---|---|---|---|---|
| Exiting approaches | Bhowmick et al. [38] | - | 84.33 % | - |
| | Basu et al. [39] | - | 80.58 % | - |
| | Bhattacharya et al. [40] | - | 95.84 % | - |
| | BHCR-CNN [41] | - | 85.96 % | - |
| | MLP (Basu et al. 2005) [42] | 96.67 % | - | - |
| | MPCA + QTLR in 2012 [43] | 98.55 % | - | - |
| | GA (Das et al. 2012B) [44] | 97.00 % | - | - |
| | SRC (Khan et al. in 2014) [45] | 94.00 % | - | - |
| | CNN+DBN [46] | 98.78 % | - | - |
| DCNN | VGG Net [19] | 97.57 | 97.56 | 96.15 |
| | All convolution [ 20] | 97.08 | 94.31 | 95.58 |
| | Network in Network (NiN) [ 21 ] | 97.36 | 96.73 | 97.24 |
| | Residual Network (ResNet) [ 22 ] | 98.51 | 97.33 | 97.64 |
| | FractalNet [ 23 ] | 98.92 | 97.87 | 97.98 |
| | DenseNet [ 24] | **99.13** | **98.31** | **98.18** |

## D. *Performance evaluation*

The testing performance is compared to several existing non-DCNN methods. The results are presented in Table III. The experimental results show that the modern DCNN models including DenseNet [24], FractalNet [23], Residual Network [22] provide better testing accuracy against the other deep learning approaches and the previously proposed classical methods. The DenseNet provides 99.13% testing accuracy for handwritten digits recognition which is the best accuracy till today. In case of 50-alphabet recognition, DenseNet yields 98.31% recognition accuracy which is almost 2.5% better than the method in [40]. To best of our knowledge, this is so far, the highest accuracy for 50 handwritten Bangla alphabets recognition. In addition, on 13 special character recognition task, DCNNs show promising recognition accuracy, especially DenseNet achieves the best accuracy which is 98.18%.

**Table IV.** Number of network parameters

| Models | Number of parameters |
|---|---|
| VGG-16 [19] | ~ 8.43 M |
| All-Conv [20] | ~ 2.26 M |
| NiN [21] | ~ 2.81 M |
| ResNet [22] | ~ 5.63 M |
| FractalNet [23] | ~ 7.84 M |
| DenseNet [24] | ~ 4.25 M |

## E. *Number of parameters*

For impartial comparison, we have trained and tested the networks with the optimized same number of parameters. Table IV shows the number of parameters used for different networks for 50-alphabet recognition. The number of network parameters for digits and special characters recognition were the same except the number of neurons in the classification layer.

## F. *Computational time for training*

We also calculate computational cost for all methods, although the computation time depends on the complexity of the architecture. Table V represents the computational time per epoch (in second) during training of all the networks for Digit-10, Alphabet-50 and SpecialChar-13 recognition task. From the Table V, it can be said that the DenseNet takes longest time during training due to its dense structure.

**Table V.** Computational time (in Sec.) per epoch for different DCNNs models on Digit-10, Alphabet-50 and SpecialChar-13 datasets.

| Models | Digit-10 | Alphabet-50 | SpecialChar-13 |
|---|---|---|---|
| VGG | 32 | 83 | 15 |
| All Conv | 7 | 23 | 4 |
| NiN | 9 | 27 | 5 |
| ResNet | 64 | 154 | 34 |
| FractalNet | 32 | 102 | 18 |
| DenseNet | 95 | 210 | 58 |

# V. CONCLUSION

Despite the advances in character recognition technology, Handwritten Bangla Characters Recognition (HBCR) has remained largely unsolved due to the presence of many confusing characters and excessive cursive that lead to low recognition accuracy. On the other hand, the deep learning has provided outstanding performance in many recognition tasks of natural language processing. In this research, we investigated handwritten Bangla characters (including digits, alphabets, and special characters) recognition approaches using different deep learning models including Visual Geometry Group (VGG) network, All convolution (All-conv), Network in Network (NiN), Residual Network (ResNet), FractalNet, and Densely Connected Network (DenseNet). The recognition accuracy of DCNN methods was also compared with the existing classical methods for HBCR. It is observed that the DenseNet provides the highest recognition accuracy in all the three experiments for digits, alphabets and special characters recognition. We have achieved recognition rate of 99.13% for Bangla handwritten digits, 98.31% for handwritten Bangla alphabet, and 98.18% for special character recognition using DenseNet which is the best recognition accuracy so far. In future, we would like to evaluate the performance of Inception Recurrent Convolutional Neural Network (IRCNN) for HBCR [47].

## Acknowledgment

Authors would like to thank the Center for Microprocessor Applications for Training Education and Research (CMATER) at the Jadavpur University, India for providing the standard CMATERdb dataset on Bangla characters.